%% file: main.tex
\documentclass[letterpaper,10pt,conference]{ieeeconf}

\IEEEoverridecommandlockouts
\input{packages}

\input{commands}

\usepackage{problems}

\title{\LARGE \bf
Distributed Certifiably Correct Range-Aided SLAM
}

\author{
Alexander~Thoms$^{1}$,
Alan~Papalia$^{2}$,
Jared~Velasquez$^{1}$,
David~M.~Rosen$^{3}$,
Sriram~Narasimhan$^{1}$
\thanks{
This work was supported by the Graduate Fellowship program at the University of California, Los Angeles, and the Chishiki-AI Fellowship program at the University of Texas at Austin.
}
\thanks{
$^{1}$Sensing and Robotics for Infrastructure Laboratory, University of California Los Angeles, Los Angeles, USA. 
{\tt\small \{adthoms, jaredvel25, snarasim\}@ucla.edu}
}
\thanks{
$^{2}$Computer Science and AI Lab (CSAIL) at the Massachusetts Institute of Technology and the MIT-WHOI Joint Program in Oceanography/Applied Ocean Science \& Engineering, Cambridge and Woods Hole, MA, USA.
{\tt\small apapalia@mit.edu}
}
\thanks{
$^{3}$Robust Autonomy Lab, Northeastern University, Boston, MA. 
{\tt\small d.rosen@northeastern.edu}
}
}

\begin{document}

\maketitle
\thispagestyle{empty}
\pagestyle{empty}

\begin{abstract}
Reliable simultaneous localization and mapping (SLAM) algorithms are necessary for safety-critical autonomous navigation. 
In the communication-constrained multi-agent setting, navigation systems increasingly use point-to-point range sensors as they afford measurements with low bandwidth requirements and known data association. The state estimation problem for these systems takes the form of range-aided (RA) SLAM. However, \textit{distributed} algorithms for solving the RA-SLAM problem lack formal guarantees on the quality of the returned estimate.
To this end, we present the \textit{first} distributed algorithm for RA-SLAM that can efficiently recover \textit{certifiably globally optimal} solutions. Our algorithm, \underline{d}istributed certifiably \underline{co}rrect \underline{RA}-SLAM (DCORA), achieves this via the Riemannian Staircase method, where computational procedures developed for distributed certifiably correct pose graph optimization are generalized to the RA-SLAM problem.
We demonstrate DCORA's efficacy on real-world multi-agent datasets by achieving absolute trajectory errors comparable to those of a state-of-the-art \textit{centralized} certifiably correct RA-SLAM algorithm. Additionally, we perform a parametric study on the structure of the RA-SLAM problem using synthetic data, revealing how common parameters affect DCORA's performance.
\end{abstract}

\section{Introduction}
\label{sec:introduction}

Realizing autonomy as a ubiquitous technology requires formal guarantees on performance. This is increasingly true for multi-agent autonomy, where coordinated planning, control, and perception must remain safety-critical for cyber-physical systems \cite{sohag2021internet} and post-disaster evaluations \cite{thoms2023graph}, among other applications. Parallel to the long-standing guarantees established by the planning \cite{russell2016artificial} and control \cite{stengel1994optimal,aastrom2021feedback} communities, a concerted effort over the past decade has been directed toward endowing perception with formal guarantees, where the prevailing algorithms underlying these methods are \textit{certifiably correct} (i.e.,\ algorithms that recover \textit{certifiably globally optimal solutions} of generally-intractable problems within a restricted but operationally relevant set of problem instances \cite{bandeira2016note,rosen2021advances}).

\begin{figure}[!t]
    \centering
    \includegraphics[width=0.5\textwidth]{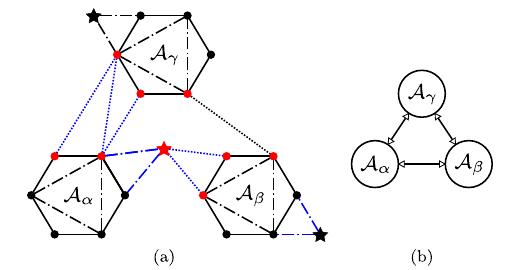}
    \vspace{-20pt}
    \caption{
    (a) Graphical representation of the RA-SLAM problem in the distributed setting. In DCORA, synchronized agents (e.g., $\mathcal{A}_\alpha, \mathcal{A}_\beta, \mathcal{A}_\gamma$) jointly estimate their state in a common frame by performing local computation steps and exchanging information over the distributed communication graph shown in (b). States, including poses (circles) and landmarks (stars), are interconnected via relative pose (black lines) and range (blue lines) measurements. To facilitate efficient communication and computation, states are labeled \textit{private} (black) or \textit{public} (red), while measurements exist as \textit{inter-agent} loop closures (dotted lines), \textit{intra-agent} loop closures (dash-dotted lines), or odometry (solid lines) as discussed in \Cref{sec:dcoras_variable_ownership_structure}.
    }
    \vspace{-20pt}
    \label{fig:dcora_graphical_representation}
\end{figure}

In this work, we advance certifiable perception for multi-agent autonomy by presenting DCORA, the \textit{first} certifiably correct algorithm for \textit{distributed} range-aided simultaneous localization and mapping (RA-SLAM). The RA-SLAM problem estimates a set of pose and landmark locations from pairwise relative pose and range measurements, and DCORA solves this problem while obeying the topology of an underlying distributed communication graph (Fig.~\ref{fig:dcora_graphical_representation}). 
Specifically, DCORA applies distributed Riemannian optimization techniques \cite{tian2021distributed} to solve the (convex) semidefinite programming (SDP) relaxation of a maximum likelihood (ML) formulation of RA-SLAM \cite{papalia2023certifiably}. In turn, solving the SDP allows DCORA to recover globally optimal RA-SLAM solutions within practically relevant noise regimes. We solve the SDP via the Riemannian Staircase method \cite{boumal2015riemannian,boumal2016non}, for which we use the \textit{Riemannian block coordinate descent} (RBCD) algorithm \cite{tian2021distributed} to identify first-order critical points (i.e., points where the Riemannian gradient of the objective function vanishes). We leverage established Karush-Kuhn-Tucker (KKT)-based conditions (which require a certain dual certificate matrix, when evaluated at the global minima, to be positive semidefinite) of the RA-SLAM problem \cite{rosen2021scalable,papalia2023certifiably} to verify global optimality, and apply distributed saddle escape techniques to descend from suboptimal critical points (if necessary). We benchmark DCORA against a state-of-the-art \textit{centralized} certifiably correct RA-SLAM algorithm on real-world datasets, demonstrating comparable absolute trajectory error (ATE) performance. Furthermore, we perform parameter sweeps across synthetic datasets to observe how common parameters (namely, the number of agents, the number of landmarks, and range measurement density) in RA-SLAM problems affect DCORA's performance in terms of convergence rate, solution precision, and SDP tightness. These contributions are summarized as follows:
\begin{enumerate}
    \item The first \textit{distributed} RA-SLAM algorithm capable of producing \textit{certifiably globally optimal estimates}.
    \item A benchmark comparison of our algorithm against a state-of-the-art \textit{centralized} certifiably correct RA-SLAM algorithm on real-world datasets.
    \item A parametric study on DCORA's convergence rate, solution quality, and SDP tightness using synthetic datasets.
    \item An open-source implementation of our algorithm\footnote{\url{https://github.com/adthoms/dcora}}.
\end{enumerate}

\section{Related Works}
\label{sec:related_works}

We briefly review works related to DCORA's method class, distributed Riemannian optimization algorithms amenable to said method class, and existing distributed RA-SLAM algorithms.

\subsection{Certifiably Correct Perception via the Riemannian Staircase}
\label{sec:certifiably_correct_perception_via_the_riemannian_staircase}

DCORA depends on a formulation of the RA-SLAM problem that allows an SDP relaxation to be constructed and subsequently solved through a series of rank-restricted SDPs of considerably lower dimension than the original SDP. Furthermore, these specific rank-restricted SDPs are equivalent to optimization over the Stiefel and Oblique manifolds, leading DCORA to instead solve a series of (smooth) manifold optimization problems. This approach -- solving an SDP via a series of lower-dimensional manifold optimization problems -- is known as the \textit{Riemannian Staircase} method \cite{boumal2015riemannian,boumal2016non}.

The first work to apply the Riemannian Staircase for certifiable perception was that of \cite{rosen2019se}, which presented SE-Sync, a certifiably correct algorithm for pose graph optimization (PGO). However, SE-Sync is an inherently centralized algorithm, which presents challenges in communication-constrained multi-agent settings. Tian \textit{et al.} \cite{tian2021distributed} addressed this with the DC2-PGO algorithm, which (to the best of our knowledge) is the only distributed certifiably correct algorithm for robotic perception. DC2-PGO presented distributed algorithms for the two key procedures of SE-Sync: optimization and certification. The capabilities of certifiable perception were then further pushed in our previous work \cite{papalia2023certifiably}, which developed a novel SDP formulation for RA-SLAM and presented CORA, a \textit{centralized} certifiably correct algorithm for RA-SLAM via the Riemannian Staircase.

\subsection{Distributed Riemannian Optimization Algorithms}
\label{sec:distributed_riemannian_optimization_algorithms}

Knuth and Barooah \cite{knuth2012collaborative,knuth2013collaborative,knuth2015distributed} presented the first distributed Riemannian optimization algorithm for robotic perception, called D-RPGO. The author's method poses PGO as a product of Stiefel manifolds, which is solved via a provably correct explicit gradient descent law. D-RPGO was developed in parallel to distributed Riemannian optimization algorithms developed for computer vision \cite{tron2009distributed,tron2014distributed}, which are also based on gradient descent. Tian \textit{et al.} \cite{tian2021distributed} developed the RBCD algorithm (and an accelerated variant RBCD++) as the core optimization procedure of DC2-PGO. RBCD/RBCD++ performs distributed block-coordinate descent over the product of (arbitrary smooth) Riemannian manifolds, and provably converges to first-order critical points with a global sublinear rate. These methods assume synchronization among agents, and Tian \textit{et al.} \cite{tian2020asynchronous} addressed this limitation by developing ASAPP, the first asynchronous and provably convergent algorithm for distributed PGO. Importantly, the authors show how ASAPP generalizes to settings where the feasible set is the product of Riemannian manifolds. Distributed Riemannian optimization for PGO was further advanced by Li \textit{et al.} \cite{li2024distributed}, who proposed combined multilevel graph partitioning (for optimization load balancing) with an accelerated Riemannian optimization method, called IRBCD. IRBCD enjoys the same convergence properties as RBCD/RBCD++ while demonstrating improvements to solution quality and convergence rate on benchmark datasets. Lastly, Fan and Murphey \cite{fan2024majorization} developed a majorization-minimization algorithm for distributed PGO that is guaranteed to converge to first-order critical points and leverages Nesterov acceleration and adaptive restart (similar to RBCD++) for accelerated convergence. We note that distributed Riemannian optimization algorithms have been developed for active robotic perception \cite{asgharivaskasi2025riemannian} and are theoretically compatible with DCORA's method class.

\subsection{Distributed RA-SLAM Algorithms}
\label{sec:distributed_ra-slam_algorithms}

Distributed SLAM algorithms have been developed to solve specific instances of the RA-SLAM problem. Such algorithms have been developed to fuse range measurements between robots \cite{xu2020decentralized,nguyen2022flexible,xu2022omni,liu2022distributed,liu2023relative} and into compound relative pose estimates between robots \cite{fishberg2022multi,cossette2022optimal,xun2023crepes,fishberg2024murp,wu2024scalable}; however, none provide a unified approach across all state and measurement combinations that arise in RA-SLAM. Moreover, these methods provide no guarantees (\textit{a priori} or a \textit{posteriori}) on the quality of the returned estimate. We note that several distributed back-end solvers \cite{cunningham2010ddf,cunningham2013ddf,matsuka2023localized,murai2024robot,murai2024distributed,mcgann2024asynchronous,mcgann2024imesa} are capable of solving the RA-SLAM problem; however, they too lack formal guarantees on solution quality. That is, these methods rely on \textit{local search} techniques (e.g., Gauss-Newton \cite{dellaert2017factor}) and are thus susceptible to significantly suboptimal critical points. Even worse, certain local search techniques (e.g., Gaussian belief propagation \cite{ortiz2021visual}) lack convergence guarantees. The limitations of existing distributed RA-SLAM algorithms and distributed back-end solvers thus necessitate a new approach for safety-critical navigation.

\section{Methodology}
\label{sec:methodology}

In this section, we first formulate the RA-SLAM problem and present its connection to Riemannian optimization. Next, we present the DCORA algorithm with a discussion on enabling efficient communication and computation via variable ownership.

\subsection{RA-SLAM Problem Formulation and the Riemannian Staircase}
\label{sec:ra-slam_problem_formulation_and_the_riemannian_staircase}

To solve the RA-SLAM problem via the Riemannian Staircase, we begin with the following ML formulation of RA-SLAM
\RaSlamMapProblem
Without loss of generality, \Cref{prob:ra-slam-map} does not distinguish landmark variables from poses (as discussed in \cite[Appendix E]{papalia2023certifiably}) to simplify mathematical presentation. To recover globally optimal solutions to \Cref{prob:ra-slam-map}, \cite{papalia2023certifiably} developed a (non-convex) quadratically constrained quadratic programming (QCQP) relaxation of \Cref{prob:ra-slam-map}, which enabled a (convex) semidefinite relaxation to be constructed via Shor's relaxation \cite{shor1987quadratic}. The QCQP relaxation is as follows
\RaSlamQcqpProblem
\Cref{prob:ra-slam-qcqp} relaxes the special orthogonal constraint $\rot_i \in \SOd$ to an orthogonality constraint, $\rot_i \in \Orthogonald$, and introduces auxiliary unit-norm vector variables, $\dist_{ij} \in \dvec$, as a reformulation of the range cost terms. From \Cref{prob:ra-slam-qcqp}, an SDP relaxation can be constructed via Shor's relaxation \cite{shor1987quadratic}
\SDPProblem
The exact steps to form this relaxation are outlined in \cite[Appendix D]{papalia2023certifiably}. Key to our approach is recognition that \Cref{prob:sdp_problem} should often admit low-rank solutions (i.e., solutions $Z^\ast$ for which $\rank Z^\ast \ll k$). Indeed, this low-rank behavior has been found in practice \cite{papalia2023certifiably}. Thus, instead of solving \Cref{prob:sdp_problem} directly via standard SDP solvers, one may instead leverage the low-rank structure by optimizing over a low-dimensional factorization $X^\top X = Z, X \in \R^{p \times k}$ for user-selectable parameter $p \ll k$. This technique, known as the Burer-Monteiro method \cite{burer2003nonlinear,burer2005local,papalia2024overview}, arrives at the following problem
\BMProblem
in which the factorization implicitly constrains the rank of the solution to be less than or equal to $p$. For our specific problem, $X$ can be structured as
\begin{equation}
    X \in \R^{p \times k } \triangleq
    \left[ 
    \rot_1 \ \dotsb \ \rot_n  \mid  r_1 \ \dotsb \ r_l  \mid t_1 \ \dotsb \ t_n
    \right].
    \label{eq:lifted_concatenated_decision_variables_for_QCQP_relaxation}
\end{equation}
For higher problem dimensions ($p > d$), variables (e.g., $\rot_1$) in \eqref{eq:lifted_concatenated_decision_variables_for_QCQP_relaxation} can be thought of as higher-dimensional generalizations of the original variables of the QCQP (\Cref{prob:ra-slam-qcqp}); \Cref{prob:bm_problem} thus represents a higher-dimensional relaxation of \Cref{prob:ra-slam-qcqp}. In fact, analysis of the constraints of \Cref{prob:bm_problem} reveals that the problem can be equivalently posed as Riemannian manifold optimization \cite{papalia2023certifiably}
\RaSlamRiemannianStairProblem
This manifold optimization representation (\Cref{prob:ra-slam-riemannian-staircase}) directly generalizes the original QCQP (\Cref{prob:ra-slam-qcqp}), as the Stiefel manifold $\St(p,d)$ generalizes the orthogonal group $\Orthogonald$ and the auxiliary variables $r_{ij}$ are still constrained to be unit-norm but can be of arbitrary dimension $p$. \Cref{prob:ra-slam-riemannian-staircase} is directly optimized in the DCORA algorithm. For notational convenience, we refer to the product manifold defined by \Cref{prob:ra-slam-riemannian-staircase}'s variables as $\ManifoldRASLAM(p,n,l) \triangleq \St(p, d)^n \times (S^{p-1})^l \times \R^{p \times n} \subset \R^{p \times k}$.

\subsection{DCORA Algorithm}
\label{sec:dcora_algorithm}

We briefly summarize the DCORA algorithm (\Cref{alg:dcora}), which is a generalization of the DC2-PGO algorithm \cite{tian2021distributed} to the RA-SLAM problem. 
\input{alg/dcora}
An initial point $X_\text{init} \in \ManifoldRASLAM(p=d,n,l)$ is obtained via distributed initialization. This initial point is \textit{lifted} to rank $p_0$: $X = Y_\text{rand} X_\text{init}$, where $Y_\text{rand} \in \St (p_0, \dimension)$ is a random Stiefel element sampled from the initial rank $p_0 \geq d$ of the Riemannian staircase.

From this lifted initial point, the Riemannian Staircase methodology is applied. First, \Cref{prob:ra-slam-riemannian-staircase} is optimized over via RBCD. Once RBCD has returned a first-order critical point $X^{\ast}$, $X^{\ast}$ is \textit{lifted} to the next rank $p+1$ in the Riemannian staircase, where a distributed verification step either certifies $X^{\ast}$ as globally optimal or constructs a second-order descent direction to descend from $X^{\ast}$. If $X^{\ast}$ is not globally optimal, then saddle point escape is performed via line search along the computed descent direction and RBCD is used to optimize from the resulting point. The distributed Riemannian Staircase algorithm continues until a certified globally optimal value $f^{\ast}_\text{SDP}$ of \Cref{prob:sdp_problem} is recovered, which is a lower bound on \Cref{prob:ra-slam-map}’s optimal value. If the SDP is tight, as it often is in practice \cite{papalia2023certifiably}, then the solution is globally optimal for \Cref{prob:ra-slam-map}. Completing the DCORA algorithm, $X^{\ast}$ is projected to the feasible set of \Cref{prob:ra-slam-map} via distributed rounding to recover pose and landmark estimates $T \in \SEd^n \triangleq \{ \rot_i \in \SOd, \tran_i \in \dvec \}_{i=1}^{\numPoses}$.

\subsection{DCORA's Variable Ownership Structure}
\label{sec:dcoras_variable_ownership_structure}

\subsubsection{Variable Partitioning}
A key notion in DCORA is that of \textit{variable ownership}. Each decision variable in \Cref{prob:ra-slam-riemannian-staircase} is owned by a single agent, and variables are partitioned by agent ownership to satisfy RBCD's input requirement. Specifically, RBCD requires a block partitioning of $X$, where a block $X_b \subseteq X$ is associated with the agent responsible for estimating those states. In DC2-PGO, agents naturally take ownership of pose variables that describe their trajectory. With the introduction of external landmarks and range measurements, variable partitioning becomes less obvious for the RA-SLAM problem. As such, we maintain DC2-PGO's pose ownership rule in DCORA and assume agents own landmarks according to a consensus step \cite{olfati2007consensus} and auxiliary unit-norm vector variables corresponding to range measurements collected by their onboard sensors.

\subsubsection{Variable Labeling}
In an agent's local search step, RBCD requires agents to maintain certain \textit{local copies} of variables belonging to other agents. To facilitate efficient sharing of these copies, poses and landmarks are labeled \textit{public} or \textit{private} according to the following definition
\begin{definition}
States that connect to the states of other agents (via pairwise measurements) are called \textit{public} states. All other states are \textit{private}. 
\end{definition}
Thus, agents sharing a pairwise measurement (referred to as an \textit{inter-agent} loop closure) need only cache the neighboring agent's public state before the local search step. If the pairwise measurement is a range measurement taken by the other agent, the corresponding auxiliary unit-norm vector variable is also cached. Information related to an agent's private states, \textit{intra-agent} loop closures (which connect non-consecutive states owned by a single agent), and odometry remains private.

\section{Experiments}
\label{sec:experiments}

Experiments conducted to evaluate DCORA are two-fold. First, we benchmark DCORA against its centralized counterpart, CORA, on public real-world multi-agent RA-SLAM datasets. Comparison is limited to CORA as (to our knowledge) this is the only other certifiably correct algorithm that can solve the general RA-SLAM problem. We refer the interested reader to \cite{papalia2023certifiably} for an in-depth comparison of CORA against the GTSAM Levenberg–Marquardt optimizer \cite{gtsam}, a popular local search back-end solver, to motivate our comparison. Second, multi-agent RA-SLAM datasets are generated by modifying public synthetic PGO datasets. These datasets are modified to assign pose variables to multiple agents, include landmarks, and establish inter- and intra-agent range measurement loop closures with controlled density. Parameter sweeps are performed based on (i) the number of agents, (ii) the number of landmarks, and (iii) range measurement density to study the RA-SLAM problem structure. DCORA's implementation for evaluation precedes the discussion of these experiments.

\subsection{Implementation for Evaluation}
\label{sec:implementation_for_evaluation}

Our implementation of DCORA centralizes two procedures from \Cref{alg:dcora} for convenience in conducting our experiments. Our experiments do not evaluate DCORA's runtime performance, and our centralized implementations only affect runtime while not affecting the performance metrics we measure. First, we employ a centralized initialization procedure in line 1 that provides quality initial estimates. Our centralized initialization procedure improves runtime over a distributed implementation by assuming all measurements and states belong to a single agent. This reduces the number of computation steps and avoids inter-agent communication when solving for $X_\text{init}$. While the solutions returned by DCORA are initialization \textit{independent}, quality initial estimates (which may be equally obtained using a distributed implementation) further improve runtime by avoiding the need to increase rank multiple times in the Riemannian Staircase to escape from spurious local minima. Second, we perform centralized minimum eigenpair calculations (as part of the solution verification step in line 7) using the centralized Lanczos-based method described in \cite{rosen2017computational}. This method has the same asymptotic convergence rate as DCORA's (and DC2-PGO's) distributed minimum eigenpair calculation, thus serving as a good proxy for the runtime performance of distributed solution verification.

\subsection{Real-World Efficacy: DCORA vs. CORA}
\label{sec:real-world_efficacy_dcora_vs_cora}

\subsubsection{Datasets}
Real-world multi-agent RA-SLAM datasets, including \texttt{TIERS} \cite{yu2023fusing} and \texttt{UTIAS MR.CLAM} \cite{leung2011utias}, are used to benchmark DCORA against CORA. To our knowledge, these are the only real-world multi-agent RA-SLAM datasets publicly available in the literature. \texttt{TIERS} comprises five ground robots with stereo cameras for visual odometry and ultra-wideband sensors to perform inter-agent ranging. One robot was stationary, acting as a landmark, while the other four moved in non-overlapping geometric patterns. \texttt{MR.CLAM} comprises five ground robots and fifteen landmarks. All robots and landmarks were equipped with barcode fiducials, and robots were equipped with cameras to obtain range (and bearing\footnote{Bearing measurements have been ignored from \texttt{MR.CLAM} to extract RA-SLAM problem instances.}) measurements to the fiducials. The robots obtained odometry from open-loop wheel velocity commands. Sessions \texttt{2}, \texttt{4}, \texttt{6}, and \texttt{7} from \texttt{MR.CLAM} are used for evaluation as the remaining sessions had substantial dropouts in ground truth data and are therefore unusable.

\subsubsection{Evaluation and Results}
For evaluation, we calculate the root mean square error (RMSE) as an ATE metric against ground truth at DCORA's and CORA's certified solutions. ATE is calculated using evo \cite{grupp2017evo} with Umeyama trajectory alignment \cite{umeyama1991least}. When calculating the minimum eigenpair in line 7 of \Cref{alg:dcora}, $S(X^{\ast})$ is regularized by a certification tolerance $\epsilon > 0$ to account for errors due to numerical precision. We assumed a default value of $\epsilon=1\text{e-3}$ and adjusted this value through the set $\{ 5\text{e-3}, 1\text{e-2}, 5\text{e-2} \}$ until DCORA and CORA were able to certify their solutions. To ensure consistency in our evaluation, we reimplemented CORA using DCORA's Riemannian trust-region back-end solver \cite{huang2018roptlib} while using the same default parameter settings as in the original CORA implementation. \Cref{tab:cora_vs_dcora_absolute_trajectory_error} summarizes ATE translation and rotation values accompanied by certification results $f^{\ast}_\text{SDP}$ and $\epsilon$.
\input{tab/cora_vs_dcora_absolute_trajectory_error}

\subsubsection{Discussion}
Results demonstrate comparable ATE performance between DCORA and CORA. This is expected, as these algorithms are (mainly) differentiated by the order of their optimization procedures (i.e., first- vs. second-order). The order of optimization does not affect the optimality of the SDP; thus, the certified solution (mathematically) should remain \textit{identical}. However, RBCD has a \textit{sublinear} convergence rate, meaning its estimates are not as precise as those provided by CORA's second-order trust-region method. As a result, DCORA requires a looser certification tolerance to verify optimality, which accounts for the small differences in $f^{\ast}_\text{SDP}$ and ATE. 

\subsection{RA-SLAM Problem Structure in the Distributed Setting}
\label{sec:ra-slam_problem_structure_in_the_distributed_setting}

\subsubsection{Datasets}
\begin{figure}[!t]
    \centering
    \includegraphics[width=0.5\textwidth]{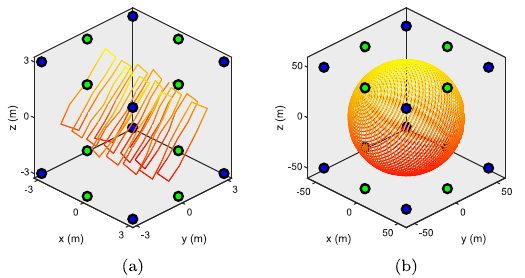}
    \vspace{-20pt}
    \caption{
    Modified (a) \texttt{smallGrid3D}\cite{tian2021distributed} and (b) \texttt{sphere2500}\cite{kaess2012isam2} pose graph optimization (PGO) datasets used to study the range-aided SLAM problem structure in the distributed setting. Original PGO datasets (with their trajectories illustrated in red) have been modified to generate 24 new datasets, which include multiple agents (not shown), landmarks (blue and green circles), and range measurements (not shown).
    }
    \vspace{-20pt}
    \label{fig:synthetic_datasets}
\end{figure}

To study the RA-SLAM problem structure in the distributed setting, two public synthetic PGO datasets, \texttt{smallGrid3D}\cite{tian2021distributed} and \texttt{sphere2500}\cite{kaess2012isam2}, are modified (Fig.~\ref{fig:synthetic_datasets}) to generate 24 multi-agent RA-SLAM problem instances.
Modified datasets evenly and contiguously distribute pose variables among two, four, or eight agents, with each dataset assuming a fixed configuration of eight or 16 landmarks. Range measurements are generated assuming a precision of $\rho_{ij} = 100 \frac{1}{\text{m}^2}$ and a sensing horizon of $r_\text{max} = 100$m. Range measurements are generated with a probability of 0.5 or 1.0 from each agent's pose variables to (i) a random agent's pose and (ii) a random landmark within a distance of $r_\text{max}$. We assume landmarks are owned by the agent with the most connecting range measurements, thus minimizing inter-agent communication.

\subsubsection{Evaluation and Results}
\begin{figure}[!t]
    \centering
    \includegraphics[width=0.5\textwidth]{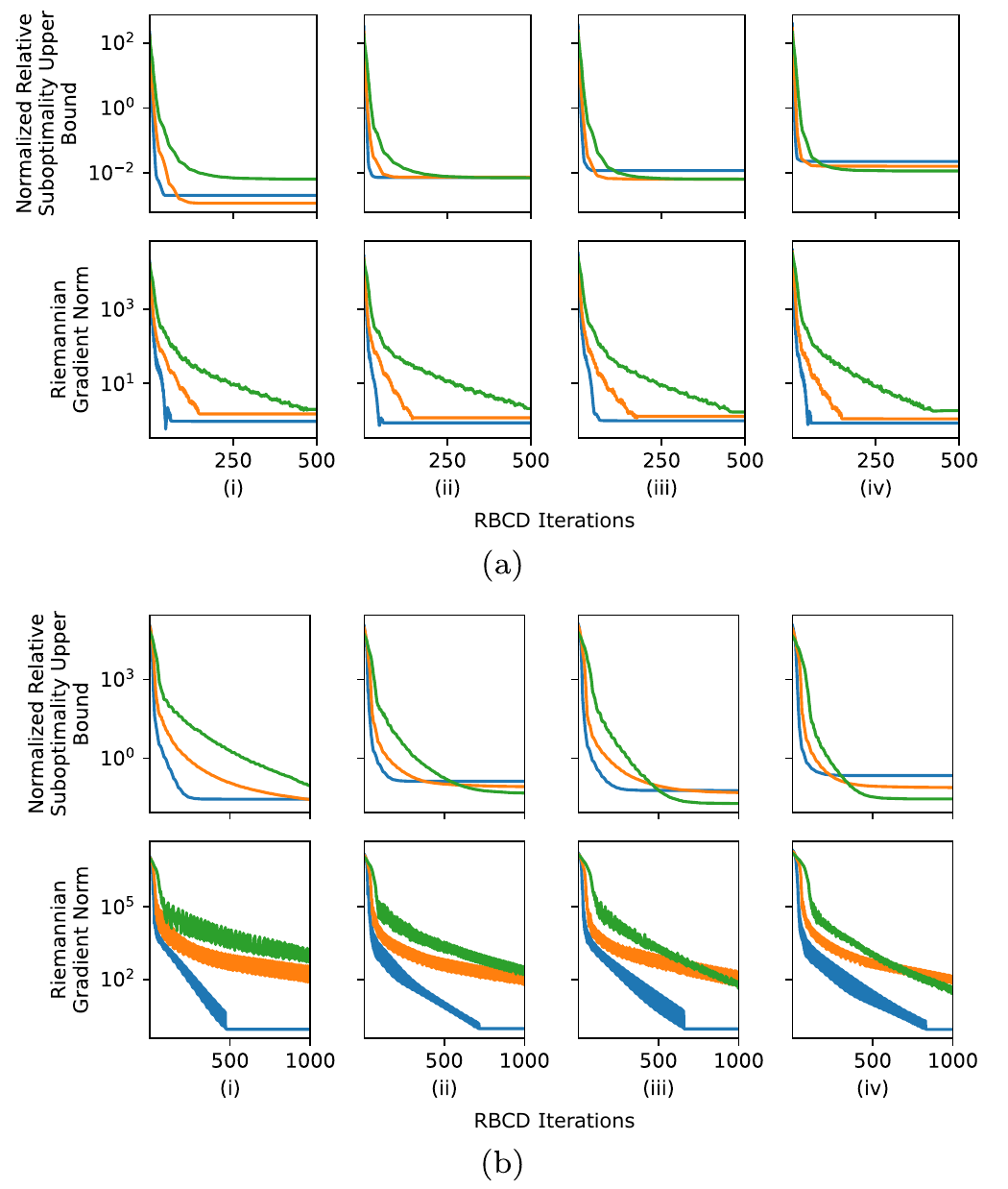}
    \vspace{-20pt}
    \caption{
    Evolution of the normalized relative suboptimality upper bound and the Riemannian gradient norm at each RBCD iterate for rank three in the Riemannian staircase. Modified datasets (a) \texttt{smallGrid3D}\cite{tian2021distributed} and (b) \texttt{sphere2500}\cite{kaess2012isam2} have the following parameters:
    (i) landmarks: 8, range measurement probability: 0.5;
    (ii) landmarks: 16, range measurement probability: 0.5;
    (iii) landmarks: 8, range measurement probability: 1.0;
    (iv) landmarks: 16, range measurement probability: 1.0;
    for two-agent (blue), four-agent (orange), and eight-agent (green) systems.
    All plots are log-linear.
    }
    \vspace{-20pt}
    \label{fig:ra_slam_problem_structure_results}
\end{figure}

For evaluation, we calculate the relative suboptimality upper bound $( f(X^k) - f^{\ast}_\text{SDP} ) / f^{\ast}_\text{SDP}$ (normalized to account for differences in scale between objective values of the modified PGO datasets) and the Riemannian gradient norm $\| \grad f(X^k) \|$ at each RBCD iterate $X^k$ for rank $p = 3$, where $f(X^k) = \langle  Q, (X^k)^{\top} X^k \rangle_F$. The former of these measures the relative suboptimality of the rounded estimates recovered from the convex relaxation, and so effectively measures \textit{how tight} our relaxation is. The latter of these measures \textit{how precisely} RBCD is able to estimate the critical points recovered from the Riemannian rank-restricted SDPs. Fig.~\ref{fig:ra_slam_problem_structure_results} illustrates the evolution of these metrics for the generated multi-agent RA-SLAM problem instances.

\subsubsection{Discussion}
As observed in Fig.~\ref{fig:ra_slam_problem_structure_results}, convergence rate and solution precision are predominantly affected by the number of agents. This trend is most clear on the modified \texttt{smallGrid3D} dataset, where fewer RBCD iterations are required to converge to a lower Riemannian gradient norm, thus suggesting faster convergence to more precisely estimated critical points. For the modified \texttt{sphere2500} dataset, an exception to this trend is observed when the range measurement probability is 1.0. Specifically, the eight-agent system demonstrates improved convergence and solution precision compared to the four-agent system. We conjecture that the increase in range measurement density compensates for the loss in connectivity in each agent's underlying pose graph, as intra-agent relative pose loop closures are redistributed as inter-agent loop closures. Observations on SDP tightness further support this conjecture, as increasing range measurement density improves tightness as the number of agents increases.

\section{Conclusion}
\label{sec:conclusion}

In this work, we presented DCORA, the first \textit{distributed certifiably correct} algorithm for range-aided simultaneous localization and mapping (RA-SLAM). DCORA generalizes the Riemannian Staircase method established for distributed certifiably correct pose graph optimization to the RA-SLAM problem and demonstrates comparable solution quality to its centralized counterpart, CORA, on real-world datasets. The structure of the RA-SLAM problem in the distributed setting was studied using parametrically varied synthetic datasets, revealing the number of agents to be the predominant factor affecting DCORA's convergence rate and solution precision. Results also showed that graph connectivity influences SDP tightness, where additional range measurements can aid solution quality for systems distributing relative pose measurements across a large number of agents. In future work, we aim to expand the real-world evaluation of DCORA across larger multi-agent systems under various noise regimes. We also plan on studying DCORA's runtime performance with a fully distributed implementation. Of broader importance to DCORA's method class, efforts to significantly enhance the speed, scalability, and communication efficiency of Riemannian block coordinate descent by incorporating (at least partial) \textit{second-}order information into its design would be a substantial contribution towards enabling efficient estimation and control methods for multi-agent systems in general. This, in addition to identifying sufficient conditions to ensure that DCORA's (and CORA's) SDP relaxation remains \textit{exact} under bounded noise for the general RA-SLAM problem, are directions for future work.

\bibliographystyle{IEEEtran}
\bibliography{bib/IEEEabrv.bib,bib/references.bib}

\end{document}

%% file: packages.tex
\usepackage{cite}
\usepackage{amsmath,amssymb,amsfonts,bm,bbm}

\usepackage{amsthm}
\usepackage{algorithm,algorithmic}
\usepackage{graphicx}
\usepackage{textcomp}
\usepackage{url}
\usepackage{xcolor}
\usepackage{hyperref}
\hypersetup{
    colorlinks=true,
    citecolor=black,
    linkcolor=blue,
    urlcolor=magenta
}
\usepackage{cleveref}
\usepackage{multirow}
\usepackage{stackengine}
\usepackage{tabularx}
\usepackage[percent]{overpic}
\usepackage{booktabs,ragged2e}
\usepackage[flushleft]{threeparttable}
\usepackage{array}
\usepackage{pifont}

%% file: commands.tex
\newcommand{\degree}[1]{{#1}\textsuperscript{$\circ$}}

\newtheorem{theorem}{Theorem}

\newtheorem{definition}[theorem]{Definition}

\def \dimension{d}
\def \liftedDimension{p}
\def \numPoses{n}
\def \numDistMeasures{l}

\def \TransformEdges {\mathcal{E}_{p}}  
\def \RangeEdges {\mathcal{E}_{r}}  

\def \dedge{(i,j)}  

\newcommand{\mat}[1]{#1}

\newcommand{\st}{\textnormal{subject to\; }}

\newcommand{\R}{\mathbb{R}}

\DeclareMathOperator{\tr}{tr}

\DeclareMathOperator{\rank}{rank}

\DeclareMathOperator{\grad}{grad}

\DeclareMathOperator{\St}{St}

\DeclareMathOperator{\Orthogonal}{O}

\DeclareMathOperator{\Orthogonald}{{\Orthogonal}(\dimension)}
\DeclareMathOperator{\SO}{SO}

\DeclareMathOperator{\SOd}{{\SO}(\dimension)}

\DeclareMathOperator{\SE}{SE}

\DeclareMathOperator{\SEd}{{\SE}(\dimension)}

\DeclareMathOperator{\Manifold}{\mathcal{M}}

\DeclareMathOperator{\ManifoldRASLAM}{\Manifold_{\text{RA-SLAM}}}

\def \tran{t}
\def \dist{r}
\def \rot{R}

\newcommand{\noisy}[1]{\tilde{#1}}

\def \ntran{\noisy{\tran}}

\def \nrot{\noisy{\rot}}
\def \ndist{\noisy{\dist}}

\def \dvec{\R^{\dimension}}
\def \dmat{\R^{\dimension \times \dimension}}

\def \matX{\mat{X}}

\def \rotOuterProd{\rot_i^\top \rot_i}

\def \sizeQcqpStateSpace{k}

\def \QcqpStateSpace{\R^{\sizeQcqpStateSpace \times \sizeQcqpStateSpace}}

%% file: alg/dcora.tex
\begin{algorithm}[!t]
\caption{ Distributed Certifiably Correct Range-Aided SLAM (DCORA) }
\label{alg:dcora}
\begin{algorithmic}[1]
\renewcommand{\algorithmicrequire}{\textbf{Input:}}
\renewcommand{\algorithmicensure}{\textbf{Output:}}
\REQUIRE Initial rank $p_0 \geq d$ for the Riemannian Staircase.
\ENSURE A feasible solution $T \in \SEd^n$ to \Cref{prob:ra-slam-map} and the lower bound $f^{\ast}_\text{SDP}$ on \Cref{prob:ra-slam-map}'s optimal value.
\STATE Obtain initial point $X_\text{init} \in \ManifoldRASLAM(p,n,l)$ for $p=d$ via distributed initialization.
\STATE Lift initial point to initial rank $X = Y_\text{rand} X_\text{init}$.
\FOR{$p = p_0, p_0+1, \dots$}
    \STATE Compute first-order critical point of \Cref{prob:bm_problem} \cite[Algorithm 3]{tian2021distributed}
    \begin{center}
        $X^{\ast} \leftarrow \text{RBCD}(X)$.
    \end{center}
    \STATE Lift to first-order critical point at next level
    \begin{center}
        $X^{\ast} \leftarrow [(X^{\ast})^{\top} \ 0]^{\top}$.
    \end{center}
    \STATE Construct corresponding dual certificate matrix \cite[Theorem 1]{papalia2023certifiably}
    \begin{center}
        $S(X^{\ast}) \triangleq Q + \sum_{i=1}^{m} \lambda_i A_i$,
    \end{center}
    where $\lambda \in \R^m \triangleq \left [ \lambda_1, \ldots, \lambda_m \right ]$ are the Lagrange multipliers at KKT point $X^{\ast}$.
    \STATE Compute minimum eigenpair \cite[Algorithm 6]{tian2021distributed}
    \begin{center}
        $(\lambda, v) \leftarrow \text{MinEig}(S(X^{\ast}))$.
    \end{center}
    \IF{$\lambda \geq 0$}
        \RETURN $X^{\ast}$.
    \ELSE
    \STATE Construct second-order descent direction
    \begin{center}
        $\dot{X}_+ \triangleq [0 \ v^{\top}]^{\top}$,
    \end{center}
    where $v \in \R^k$ is the eigenvector corresponding to the smallest eigenvalue of $S(X^{\ast})$ which satisfies $v^{\top} S(X^{\ast}) v < 0$.
    \STATE Descend from $X^{\ast}$ \cite[Algorithm 7]{tian2021distributed}
    \begin{center}
        $X \leftarrow \text{EscapeSaddle}(X^{\ast}, \dot{X}_+)$.
    \end{center}
    \ENDIF
\ENDFOR
\STATE Recover the optimal value of the SDP relaxation
\begin{center}
$f^{\ast}_\text{SDP} = \langle  Q, {X^{\ast}}^{\top} X^{\ast} \rangle_F$,
\end{center}
where $\langle \cdot, \cdot \rangle_F$ is the Frobenius inner product.
\STATE Project $X^{\ast}$ to the feasible set of \Cref{prob:ra-slam-map} \cite[Algorithm 3]{papalia2023certifiably} via distributed rounding to obtain $T \in \SEd^n$.
\RETURN $T, f^{\ast}_\text{SDP}$.
\end{algorithmic}
\end{algorithm}

%% file: tab/cora_vs_dcora_absolute_trajectory_error.tex
\begin{table}[!t]
\renewcommand{\arraystretch}{1.0}
\caption{DCORA vs. CORA on Public Real-World Multi-Agent RA-SLAM Datasets}
\label{tab:cora_vs_dcora_absolute_trajectory_error}

\begin{threeparttable}
    \setlength{\tabcolsep}{5.5pt}
    
    \begin{tabularx}{\columnwidth}{cc cc cc}
    \toprule
    \multirow{2}{*}{\textbf{Dataset\tnote{$\dagger$}}} & 
    \multirow{2}{*}{\textbf{Algorithm}} & 
    \multicolumn{2}{c}{\textbf{Certification\tnote{$\ddagger$}}} &
    \multicolumn{2}{c}{\textbf{ATE (RMSE)\tnote{$\ast$}}} \\
    & 
    & \textbf{$f^{\ast}_{\text{SDP}}$}
    & \textbf{$\epsilon$}
    & \textbf{Trans. [m]}
    & \textbf{Rot. [\degree{}]} \\
    \midrule    
    \multirow{2}{*}{\texttt{TIERS} }
    & DCORA & 1920.4 & 1e-2 & 0.04 & 0.05 \\
    & CORA  & 1921.9 & 1e-3 & 0.04 & 0.05 \\
    \midrule
    \multirow{2}{*}{\texttt{MR.CLAM 2} } 
    & DCORA & 7032.9 & 5e-3 & \textbf{0.15} & 0.12 \\
    & CORA  & 7034.6 & 1e-3 & 0.17 & 0.12 \\
    \midrule
    \multirow{2}{*}{\texttt{MR.CLAM 4} }
    & DCORA & 4173.2 & 5e-3 & 0.13 & 0.10 \\
    & CORA  & 4173.1 & 1e-3 & 0.13 & 0.10 \\
    \midrule
    \multirow{2}{*}{\texttt{MR.CLAM 6} }
    & DCORA & 3124.8 & 5e-3 & 0.27 & 0.14 \\
    & CORA  & 3124.5 & 1e-3 & \textbf{0.26} & 0.14 \\
    \midrule
    \multirow{2}{*}{\texttt{MR.CLAM 7} }
    & DCORA & 3091.4 & 5e-2 & 0.31 & 0.14 \\
    & CORA  & 3038.4 & 1e-2 & \textbf{0.27} & \textbf{0.13} \\
    \bottomrule
    \end{tabularx}

    \smallskip
    \scriptsize
    \begin{tablenotes}
    \RaggedRight
    \item[$\dagger$] Datasets include \texttt{TIERS}\cite{yu2023fusing} and sessions \texttt{2}, \texttt{4}, \texttt{6}, and \texttt{7} of \texttt{MR.CLAM}\cite{leung2011utias}.
    \item[$\ddagger$] All solutions are certified.
    \item[$\ast$] Bold values indicate best performance.
    \end{tablenotes}
    \end{threeparttable}
\vspace{-20pt}
\end{table}

%% file: main.bbl
\begin{thebibliography}{10}
\providecommand{\url}[1]{#1}
\csname url@samestyle\endcsname
\providecommand{\newblock}{\relax}
\providecommand{\bibinfo}[2]{#2}
\providecommand{\BIBentrySTDinterwordspacing}{\spaceskip=0pt\relax}
\providecommand{\BIBentryALTinterwordstretchfactor}{4}
\providecommand{\BIBentryALTinterwordspacing}{\spaceskip=\fontdimen2\font plus
\BIBentryALTinterwordstretchfactor\fontdimen3\font minus
  \fontdimen4\font\relax}
\providecommand{\BIBforeignlanguage}[2]{{%
\expandafter\ifx\csname l@#1\endcsname\relax
\typeout{** WARNING: IEEEtran.bst: No hyphenation pattern has been}%
\typeout{** loaded for the language `#1'. Using the pattern for}%
\typeout{** the default language instead.}%
\else
\language=\csname l@#1\endcsname
\fi
#2}}
\providecommand{\BIBdecl}{\relax}
\BIBdecl

\bibitem{sohag2021internet}
S.~Kabir, ``Internet of things and safety assurance of cooperative
  cyber-physical systems: Opportunities and challenges,'' \emph{IEEE Internet
  of Things Magazine}, vol.~4, no.~2, pp. 74--78, 2021.

\bibitem{thoms2023graph}
A.~Thoms, Z.~Al-Sabbag, and S.~Narasimhan,
  ``\BIBforeignlanguage{eng}{Graph-based structural joint pose estimation in
  non-line-of-sight conditions},'' \emph{\BIBforeignlanguage{eng}{Earthquake
  Engineering and Engineering Vibration}}, vol.~22, no.~2, pp. 371--386, 2023.

\bibitem{russell2016artificial}
S.~J. Russell, P.~Norvig, and M.-W. Chang,
  \emph{\BIBforeignlanguage{eng}{Artificial intelligence a modern approach}},
  fourth edition, global edition.~ed.\hskip 1em plus 0.5em minus 0.4em\relax
  Harlow, England: Pearson, 2022.

\bibitem{stengel1994optimal}
R.~F. Stengel, \emph{\BIBforeignlanguage{eng}{Optimal control and estimation /
  Robert F. Stengel.}}, ser. Dover books on mathematics.\hskip 1em plus 0.5em
  minus 0.4em\relax New York: Dover Publications, 1994 - 1986.

\bibitem{aastrom2021feedback}
\emph{\BIBforeignlanguage{eng}{Feedback Systems - An Introduction for
  Scientists and Engineers}}.\hskip 1em plus 0.5em minus 0.4em\relax Princeton
  University Press, 2021.

\bibitem{bandeira2016note}
\BIBentryALTinterwordspacing
A.~S. Bandeira, ``A note on probably certifiably correct algorithms,''
  \emph{Comptes Rendus Mathematique}, vol. 354, no.~3, pp. 329--333, 2016.
  [Online]. Available:
  \url{https://www.sciencedirect.com/science/article/pii/S1631073X15003519}
\BIBentrySTDinterwordspacing

\bibitem{rosen2021advances}
D.~M. Rosen, K.~J. Doherty, A.~Terán~Espinoza, and J.~J. Leonard,
  ``\BIBforeignlanguage{eng}{Advances in inference and representation for
  simultaneous localization and mapping},''
  \emph{\BIBforeignlanguage{eng}{Annual review of control, robotics, and
  autonomous systems}}, vol.~4, no.~1, pp. 215--242, 2021.

\bibitem{tian2021distributed}
Y.~Tian, K.~Khosoussi, D.~M. Rosen, and J.~P. How,
  ``\BIBforeignlanguage{eng}{Distributed certifiably correct pose-graph
  optimization},'' \emph{\BIBforeignlanguage{eng}{IEEE transactions on
  robotics}}, vol.~37, no.~6, pp. 2137--2156, 2021.

\bibitem{papalia2023certifiably}
A.~Papalia, A.~Fishberg, B.~W. O'Neill, J.~P. How, D.~M. Rosen, and J.~J.
  Leonard, ``Certifiably correct range-aided slam,'' \emph{IEEE Transactions on
  Robotics}, pp. 1--20, 2024.

\bibitem{boumal2015riemannian}
N.~Boumal, ``A riemannian low-rank method for optimization over semidefinite
  matrices with block-diagonal constraints,'' \emph{arXiv preprint
  arXiv:1506.00575}, 2015.

\bibitem{boumal2016non}
N.~Boumal, V.~Voroninski, and A.~Bandeira, ``The non-convex burer-monteiro
  approach works on smooth semidefinite programs,'' \emph{Advances in Neural
  Information Processing Systems}, vol.~29, 2016.

\bibitem{rosen2021scalable}
D.~M. Rosen, ``\BIBforeignlanguage{eng}{Scalable low-rank semidefinite
  programming for certifiably correct machine perception},'' in
  \emph{\BIBforeignlanguage{eng}{Algorithmic Foundations of Robotics XIV}},
  ser. Springer Proceedings in Advanced Robotics.\hskip 1em plus 0.5em minus
  0.4em\relax Cham: Springer International Publishing, 2021, pp. 551--566.

\bibitem{rosen2019se}
D.~M. Rosen, L.~Carlone, A.~S. Bandeira, and J.~J. Leonard,
  ``\BIBforeignlanguage{eng}{Se-sync: A certifiably correct algorithm for
  synchronization over the special euclidean group},''
  \emph{\BIBforeignlanguage{eng}{The International journal of robotics
  research}}, vol.~38, no. 2-3, pp. 95--125, 2019.

\bibitem{knuth2012collaborative}
J.~Knuth and P.~Barooah, ``Collaborative 3d localization of robots from
  relative pose measurements using gradient descent on manifolds,'' in
  \emph{2012 IEEE International Conference on Robotics and Automation}, 2012,
  pp. 1101--1106.

\bibitem{knuth2013collaborative}
------, ``Collaborative localization with heterogeneous inter-robot
  measurements by riemannian optimization,'' in \emph{2013 IEEE International
  Conference on Robotics and Automation}, 2013, pp. 1534--1539.

\bibitem{knuth2015distributed}
------, ``Distributed collaborative 3d pose estimation of robots from
  heterogeneous relative measurements: an optimization on manifold approach,''
  \emph{Robotica}, vol.~33, no.~7, p. 1507–1535, 2015.

\bibitem{tron2009distributed}
R.~Tron and R.~Vidal, ``Distributed image-based 3-d localization of camera
  sensor networks,'' in \emph{Proceedings of the 48h IEEE Conference on
  Decision and Control (CDC) held jointly with 2009 28th Chinese Control
  Conference}, 2009, pp. 901--908.

\bibitem{tron2014distributed}
------, ``Distributed 3-d localization of camera sensor networks from 2-d image
  measurements,'' \emph{IEEE Transactions on Automatic Control}, vol.~59,
  no.~12, pp. 3325--3340, 2014.

\bibitem{tian2020asynchronous}
Y.~Tian, A.~Koppel, A.~S. Bedi, and J.~P. How, ``Asynchronous and parallel
  distributed pose graph optimization,'' \emph{IEEE Robotics and Automation
  Letters}, vol.~5, no.~4, pp. 5819--5826, 2020.

\bibitem{li2024distributed}
C.~Li, G.~Guo, P.~Yi, and Y.~Hong, ``Distributed pose-graph optimization with
  multi-level partitioning for multi-robot slam,'' \emph{IEEE Robotics and
  Automation Letters}, vol.~9, no.~6, pp. 4926--4933, 2024.

\bibitem{fan2024majorization}
T.~Fan and T.~D. Murphey, ``Majorization minimization methods for distributed
  pose graph optimization,'' \emph{IEEE Transactions on Robotics}, vol.~40, pp.
  22--42, 2024.

\bibitem{asgharivaskasi2025riemannian}
A.~Asgharivaskasi, F.~Girke, and N.~Atanasov, ``Riemannian optimization for
  active mapping with robot teams,'' \emph{IEEE Transactions on Robotics},
  vol.~41, pp. 1077--1097, 2025.

\bibitem{xu2020decentralized}
H.~Xu, L.~Wang, Y.~Zhang, K.~Qiu, and S.~Shen, ``Decentralized
  visual-inertial-uwb fusion for relative state estimation of aerial swarm,''
  in \emph{2020 IEEE International Conference on Robotics and Automation
  (ICRA)}, 2020, pp. 8776--8782.

\bibitem{nguyen2022flexible}
T.~H. Nguyen, T.-M. Nguyen, and L.~Xie, ``Flexible and resource-efficient
  multi-robot collaborative visual-inertial-range localization,'' \emph{IEEE
  Robotics and Automation Letters}, vol.~7, no.~2, pp. 928--935, 2022.

\bibitem{xu2022omni}
H.~Xu, Y.~Zhang, B.~Zhou, L.~Wang, X.~Yao, G.~Meng, and S.~Shen, ``Omni-swarm:
  A decentralized omnidirectional visual–inertial–uwb state estimation
  system for aerial swarms,'' \emph{IEEE Transactions on Robotics}, vol.~38,
  no.~6, pp. 3374--3394, 2022.

\bibitem{liu2022distributed}
R.~Liu, Z.~Deng, Z.~Cao, M.~Shalihan, B.~P.~L. Lau, K.~Chen, K.~Bhowmik,
  C.~Yuen, and U.-X. Tan, ``Distributed ranging slam for multiple robots with
  ultra-wideband and odometry measurements,'' in \emph{2022 IEEE/RSJ
  International Conference on Intelligent Robots and Systems (IROS)}, 2022, pp.
  13\,684--13\,691.

\bibitem{liu2023relative}
J.~Liu and G.~Hu, ``Relative localization estimation for multiple robots via
  the rotating ultra-wideband tag,'' \emph{IEEE Robotics and Automation
  Letters}, vol.~8, no.~7, pp. 4187--4194, 2023.

\bibitem{fishberg2022multi}
A.~Fishberg and J.~P. How, ``Multi-agent relative pose estimation with uwb and
  constrained communications,'' in \emph{2022 IEEE/RSJ International Conference
  on Intelligent Robots and Systems (IROS)}, 2022, pp. 778--785.

\bibitem{cossette2022optimal}
C.~C. Cossette, M.~A. Shalaby, D.~Saussié, J.~L. Ny, and J.~R. Forbes,
  ``Optimal multi-robot formations for relative pose estimation using range
  measurements,'' in \emph{2022 IEEE/RSJ International Conference on
  Intelligent Robots and Systems (IROS)}, 2022, pp. 2431--2437.

\bibitem{xun2023crepes}
Z.~Xun, J.~Huang, Z.~Li, Z.~Ying, Y.~Wang, C.~Xu, F.~Gao, and Y.~Cao, ``Crepes:
  Cooperative relative pose estimation system,'' in \emph{2023 IEEE/RSJ
  International Conference on Intelligent Robots and Systems (IROS)}, 2023, pp.
  5274--5281.

\bibitem{fishberg2024murp}
A.~Fishberg, B.~J. Quiter, and J.~P. How, ``Murp: Multi-agent ultra-wideband
  relative pose estimation with constrained communications in 3d
  environments,'' \emph{IEEE Robotics and Automation Letters}, vol.~9, no.~11,
  pp. 10\,612--10\,619, 2024.

\bibitem{wu2024scalable}
T.~Wu and F.~Gao, ``{Scalable Distance-based Multi-Agent Relative State
  Estimation via Block Multiconvex Optimization},'' in \emph{Proceedings of
  Robotics: Science and Systems}, Delft, Netherlands, July 2024.

\bibitem{cunningham2010ddf}
A.~Cunningham, M.~Paluri, and F.~Dellaert, ``\BIBforeignlanguage{eng}{Ddf-sam:
  Fully distributed slam using constrained factor graphs},'' in
  \emph{\BIBforeignlanguage{eng}{2010 IEEE/RSJ International Conference on
  Intelligent Robots and Systems}}.\hskip 1em plus 0.5em minus 0.4em\relax
  IEEE, 2010, pp. 3025--3030.

\bibitem{cunningham2013ddf}
A.~Cunningham, V.~Indelman, and F.~Dellaert, ``Ddf-sam 2.0: Consistent
  distributed smoothing and mapping,'' in \emph{2013 IEEE International
  Conference on Robotics and Automation}, 2013, pp. 5220--5227.

\bibitem{matsuka2023localized}
K.~Matsuka and S.-J. Chung, ``Localized and incremental probabilistic inference
  for large-scale networked dynamical systems,'' \emph{IEEE Transactions on
  Robotics}, vol.~39, no.~5, pp. 3516--3535, 2023.

\bibitem{murai2024robot}
R.~Murai, J.~Ortiz, S.~Saeedi, P.~H.~J. Kelly, and A.~J. Davison, ``A robot web
  for distributed many-device localization,'' \emph{IEEE Transactions on
  Robotics}, vol.~40, pp. 121--138, 2024.

\bibitem{murai2024distributed}
R.~Murai, I.~Alzugaray, P.~H. Kelly, and A.~J. Davison, ``Distributed
  simultaneous localisation and auto-calibration using gaussian belief
  propagation,'' \emph{IEEE Robotics and Automation Letters}, vol.~9, no.~3,
  pp. 2136--2143, 2024.

\bibitem{mcgann2024asynchronous}
D.~McGann, K.~Lassak, and M.~Kaess, ``Asynchronous distributed smoothing and
  mapping via on-manifold consensus admm,'' in \emph{2024 IEEE International
  Conference on Robotics and Automation (ICRA)}, 2024, pp. 4577--4583.

\bibitem{mcgann2024imesa}
D.~McGann and M.~Kaess, ``{iMESA: Incremental Distributed Optimization for
  Collaborative Simultaneous Localization and Mapping},'' in \emph{Proceedings
  of Robotics: Science and Systems}, Delft, Netherlands, July 2024.

\bibitem{dellaert2017factor}
F.~Dellaert, M.~Kaess \emph{et~al.}, ``Factor graphs for robot perception,''
  \emph{Foundations and Trends{\textregistered} in Robotics}, vol.~6, no. 1-2,
  pp. 1--139, 2017.

\bibitem{ortiz2021visual}
J.~Ortiz, T.~Evans, and A.~J. Davison, ``A visual introduction to gaussian
  belief propagation,'' \emph{arXiv preprint arXiv:2107.02308}, 2021.

\bibitem{shor1987quadratic}
N.~Z. Shor, ``Quadratic optimization problems,'' \emph{Soviet Journal of
  Computer and Systems Sciences}, vol.~25, pp. 1--11, 1987.

\bibitem{burer2003nonlinear}
S.~Burer and R.~D. Monteiro, ``A nonlinear programming algorithm for solving
  semidefinite programs via low-rank factorization,'' \emph{Mathematical
  programming}, vol.~95, no.~2, pp. 329--357, 2003.

\bibitem{burer2005local}
------, ``Local minima and convergence in low-rank semidefinite programming,''
  \emph{Mathematical programming}, vol. 103, no.~3, pp. 427--444, 2005.

\bibitem{papalia2024overview}
A.~Papalia, Y.~Tian, D.~M. Rosen, J.~P. How, and J.~J. Leonard, ``An overview
  of the burer-monteiro method for certifiable robot perception,'' \emph{arXiv
  preprint arXiv:2410.00117}, 2024.

\bibitem{boumal2023book}
N.~Boumal, \emph{An introduction to optimization on smooth manifolds}.\hskip
  1em plus 0.5em minus 0.4em\relax Cambridge University Press, 2023.

\bibitem{olfati2007consensus}
R.~Olfati-Saber, J.~A. Fax, and R.~M. Murray, ``Consensus and cooperation in
  networked multi-agent systems,'' \emph{Proceedings of the IEEE}, vol.~95,
  no.~1, pp. 215--233, 2007.

\bibitem{gtsam}
\BIBentryALTinterwordspacing
F.~Dellaert and G.~Contributors, ``borglab/gtsam,'' May 2022. [Online].
  Available: \url{https://github.com/borglab/gtsam}
\BIBentrySTDinterwordspacing

\bibitem{rosen2017computational}
D.~Rosen and L.~Carlone, ``Computational enhancements for certifiably correct
  slam,'' in \emph{IEEE/RSJ International Conference on Intelligent Robots and
  Systems (IROS)}, 2017.

\bibitem{yu2023fusing}
X.~Yu, I.~Catalano, P.~T. Mor{\'o}n, S.~Salimpour, T.~Westerlund, and J.~P.
  Queralta, ``Fusing odometry, uwb ranging, and spatial detections for relative
  multi-robot localization,'' \emph{arXiv preprint arXiv:2304.06264}, 2023.

\bibitem{leung2011utias}
\BIBentryALTinterwordspacing
K.~Y. Leung, Y.~Halpern, T.~D. Barfoot, and H.~H. Liu, ``The utias multi-robot
  cooperative localization and mapping dataset,'' \emph{The International
  Journal of Robotics Research}, vol.~30, no.~8, pp. 969--974, 2011. [Online].
  Available: \url{https://doi.org/10.1177/0278364911398404}
\BIBentrySTDinterwordspacing

\bibitem{grupp2017evo}
M.~Grupp, ``evo: Python package for the evaluation of odometry and slam.''
  \url{https://github.com/MichaelGrupp/evo}, 2017.

\bibitem{umeyama1991least}
S.~Umeyama, ``\BIBforeignlanguage{eng}{Least-squares estimation of
  transformation parameters between two point patterns},''
  \emph{\BIBforeignlanguage{eng}{IEEE Transactions on Pattern Analysis and
  Machine Intelligence}}, vol.~13, no.~4, pp. 376--380, 1991.

\bibitem{huang2018roptlib}
\BIBentryALTinterwordspacing
W.~Huang, P.-A. Absil, K.~A. Gallivan, and P.~Hand, ``Roptlib: An
  object-oriented c++ library for optimization on riemannian manifolds,''
  \emph{ACM Trans. Math. Softw.}, vol.~44, no.~4, Jul. 2018. [Online].
  Available: \url{https://doi.org/10.1145/3218822}
\BIBentrySTDinterwordspacing

\bibitem{kaess2012isam2}
\BIBentryALTinterwordspacing
M.~Kaess, H.~Johannsson, R.~Roberts, V.~Ila, J.~J. Leonard, and F.~Dellaert,
  ``isam2: Incremental smoothing and mapping using the bayes tree,'' \emph{The
  International Journal of Robotics Research}, vol.~31, no.~2, pp. 216--235,
  2012. [Online]. Available: \url{https://doi.org/10.1177/0278364911430419}
\BIBentrySTDinterwordspacing

\end{thebibliography}
